\documentclass{article}

\usepackage[preprint]{neurips_2025}

\usepackage[utf8]{inputenc} 
\usepackage[T1]{fontenc}    
\usepackage{hyperref}       
\usepackage{url}            
\usepackage{booktabs}       
\usepackage{amsfonts}       
\usepackage{nicefrac}       
\usepackage{microtype}      
\usepackage{xcolor}         
\usepackage{acronym}        
\usepackage{graphicx}
\usepackage{amsmath}
\usepackage{algorithm}
\usepackage{algpseudocode}
\usepackage{subcaption}     
\usepackage{enumitem}     

\acrodef{cnn}[CNN]{convolutional neural networks}
\acrodef{dwi}[DWI]{diffusion-weighted imaging}
\acrodef{dmri}[dMRI]{diffusion magnetic resonance imaging}
\acrodef{dti}[DTI]{diffusion tensor imaging}
\acrodef{qbi}[QBI]{Q-ball imaging}
\acrodef{fodf}[fODF]{fiber orientation distribution function}
\acrodef{rf}[RF]{random forest}
\acrodef{eof}[EoF]{end of fiber}
\acrodef{rnn}[RNN]{recurrent neural network}
\acrodef{cnn3d}[3D-CNN]{3D convolutional neural network}
\acrodef{ffn}[FFN]{feed-forward network}
\acrodef{fa}[FA]{fractional anisotropy}
\acrodef{kld}[KL-Div]{Kullback-Leibler divergence}
\acrodef{vc}[VC]{valid connection}
\acrodef{ol}[OL]{overlap}
\acrodef{or}[OR]{overreach}

\title{TractoTransformer: Diffusion MRI Streamline Tractography using CNN and Transformer Networks}

\author{%
  Itzik Waizman \\
  Ben-Gurion University of the Negev\\
  Be'er-Sheva, Israel \\
  \texttt{itzikwei@post.bgu.ac.il} \\
  \And
  Yakov Gusakov \\
  Ben-Gurion University of the Negev \\
  Be'er-Sheva, Israel \\
  \texttt{gusakovy@post.bgu.ac.il} \\
  \And
  Itay Benou \\
  Ben-Gurion University of the Negev \\
  Be'er-Sheva, Israel \\
  \texttt{benoui@post.bgu.ac.il} \\
  \And
  Tammy Riklin Raviv \\
  Ben-Gurion University of the Negev \\
  Be'er-Sheva, Israel \\
  \texttt{rrtammy@bgu.ac.il} \\
}

\begin{document}

\maketitle

\begin{abstract}
White matter tractography is an advanced neuroimaging technique that reconstructs the 3D white matter pathways of the brain from diffusion MRI data. It can be framed as a pathfinding problem aiming to infer neural fiber trajectories from noisy and ambiguous measurements, facing challenges such as crossing, merging, and fanning white-matter configurations.
In this paper, we propose a novel tractography method that leverages Transformers to model the sequential nature of white matter streamlines, enabling the prediction of fiber directions by integrating both the trajectory context and current diffusion MRI measurements. To incorporate spatial information, we utilize CNNs that extract microstructural features from local neighborhoods around each voxel. By combining these complementary sources of information, our approach improves the precision and completeness of neural pathway mapping compared to traditional tractography models. We evaluate our method with the Tractometer toolkit, achieving competitive performance against state-of-the-art approaches, and present qualitative results on the TractoInferno dataset, demonstrating strong generalization to real-world data.
\end{abstract}


\section{Introduction}
\label{intro}
Tractography is a key technique for analyzing \ac{dwi} data, aiming to reconstruct the complex 3D trajectories of white matter fibers—a fundamental step in understanding brain connectivity, development and neurological disorders \cite{FIBERSTUDY, DISORDER}. It exploits the principle that water molecules preferentially diffuse along axonal fibers, enabling the indirect estimation of fiber orientations from diffusion-weighted measurements acquired via \ac{dmri}. Conceptually, tractography can be framed as a pathfinding problem: inferring plausible neural fiber pathways from noisy and ambiguous data while addressing challenges such as crossing, merging, and fanning fiber bundles. Traditional tractography methods rely on mathematical models that fit an estimated \ac{fodf} to the measured \ac{dwi} at each voxel, such as \ac{dti} \cite{dti}, multi-tensor models~\cite{caan2010estimation}, ball-and-sticks~\cite{behrens2003characterization}, \ac{qbi} \cite{qball}, and spherical deconvolution \cite{sd}. These orientation functions serve as local directional priors that guide the reconstruction of white matter pathways using deterministic, probabilistic, or combinatorial tracking strategies.
 
While classical tractography methods have significantly advanced our understanding of white matter architecture, they remain constrained by model-based assumptions—such as simplified representations of diffusion and voxel-wise independence \cite{LIMIT}. These limitations have spurred the development of data-driven alternatives that learn directly from \ac{dmri} data \cite{POULIN201937}. Machine learning approaches offer greater flexibility in capturing complex white matter configurations, including fiber crossings and branchings, without imposing explicit assumptions about tissue properties or the \ac{dmri} signal.

Although recent learning-based strategies have shown encouraging results, many still fall short of fully exploiting the underlying structure of the diffusion measurements, as they predict each voxel’s orientation in isolation—disregarding either spatial dependencies~\cite{RF, NEHER2017, LearnToTrack, DEEPTRACT, Entrack} or the sequential structure of white matter tracts~\cite{koppers2016direct, wasserthal2018tract, reisert2018hamlet, nath2019deep, sedlar2021diffusion, li2021superdti}. Consequently, fiber orientation predictions tend to degrade in anatomically intricate or ambiguous regions.

In this work, we effectively leverage both the spatial and contextual information inherent in the data by proposing a spatio-sequential formulation of the \ac{fodf} estimation task. Specifically, local features are first extracted from the \ac{dmri} volume using a 3D CNN, then passed to a decoder-only Transformer that predicts the \ac{fodf} at each point along a streamline, conditioned on the preceding trajectory—offering a principled integration of fiber orientation features. Our contributions include:
\begin{itemize}[leftmargin=18pt]
\item An algorithmic formulation of tractography as a pathfinding task, inspired by attention-based auto-regressive language models and spatially aware encoding.
    \item A tractography model that achieves state-of-the-art performance on a widely used benchmark, outperforming existing methods in key metrics.
    \item Open-source code infrastructure for training tractography models on multi-subject datasets, with support for in vivo diffusion MRI scans.
\end{itemize}


\section{Related Work}
\label{relwork}
In recent years, machine learning has emerged as a powerful tool for advancing tractography, moving beyond the limitations of traditional model-based approaches~\cite{NEHER2025315}. Early work by Neher et al. (2015, 2017)~\cite{RF,NEHER2017} introduced a pioneering machine learning-based tractography method that uses a \ac{rf} classifier to guide streamline progression based on raw diffusion MRI data. This method demonstrated improved performance, particularly in complex fiber configurations, by taking advantage of data-driven decision-making to predict fiber directions and terminations.

Building on the idea of sequential data processing, Poulin et al. (2017) proposed LearnToTrack \cite{LearnToTrack} and Benou et al. (2019) proposed DeepTract~\cite{DEEPTRACT}. Both frameworks utilize \acp{rnn} for tractography, but differ in the way they frame the task. The former addressed streamline tractography as a regression problem by predicting continuous (deterministic) tracking directions, while the latter takes a classification approach by outputting a distribution over discrete directions on the unit sphere, thus allowing probabilistic tractography as well as deterministic.
By treating streamlines as sequences of \ac{dwi} data, \ac{rnn} models capture the sequential dependencies of the data as context for inferring local fiber orientations. While \acp{rnn} enable sequential data processing, they are now often outperformed by Transformers, which offer better parallelization and long-range dependency handling.

Wegmayr et al. (2021) introduced Entrack \cite{Entrack}, a probabilistic spherical regression approach that incorporates entropy regularization to manage uncertainty in fiber orientation estimation. Entrack uses the Fisher-von-Mises distribution to model the posterior distribution of local streamline directions, enhancing the robustness of the tractography in noisy conditions. This probabilistic approach is particularly well-suited for complex fiber architectures where multiple crossing fibers are present.

The exploration of reinforcement learning for tractography was advanced by Théberge et al. (2021) with the introduction of TrackToLearn \cite{TrackToLearn}. This framework frames tractography as a reinforcement learning problem, where an agent learns to navigate white matter pathways by optimizing a reward function based on alignment with principal diffusion directions. This method does not require ground-truth tractograms for training, making it versatile across different datasets.

Hosseini et al. (2022) proposed CTtrack \cite{CTtrack}, a method combining CNNs and Transformers for \ac{fodf} estimation. In CTtrack, a CNN projects diffusion MRI data to a lower-dimensional space, which is then processed by a Transformer to estimate \acp{fodf} as spherical harmonic coefficients. 
While both CTtrack and our proposed TractoTransformer combine CNNs and Transformers, their modeling paradigms differ fundamentally. CTtrack processes \ac{dwi} data in a non-sequential manner, while our proposed TractoTransformer treats tractography as an auto-regressive sequence modeling task, offering a more structured and context-aware framework tailored to the intrinsic sequential nature of tractography.


\section{Methodology}
\label{method}
The main goal of the proposed TractoTransformer method is to extract streamlines—sequences of \((x, y, z)\) coordinates representing fiber pathways—from volumetric \ac{dwi} data. The core concepts are illustrated in Figure~\ref{fig:nn-diagram}. A detailed formulation and description of the data are provided in Section~\ref{formulation}. The network architecture and its key components are presented in Section~\ref{arch}, while Section~\ref{optim} outlines the training process for conditional \acp{fodf} prediction and the optimization strategies used to enhance model performance. Finally, Section~\ref{inference} describes the inference procedure for streamline tracking on unseen data using the trained model.
\subsection{Model Formulation}
\label{formulation}
\begin{figure}[htbp]
    \centering
    \begin{subfigure}[b]{0.15\textwidth}
        \centering
        \includegraphics[width=\textwidth]{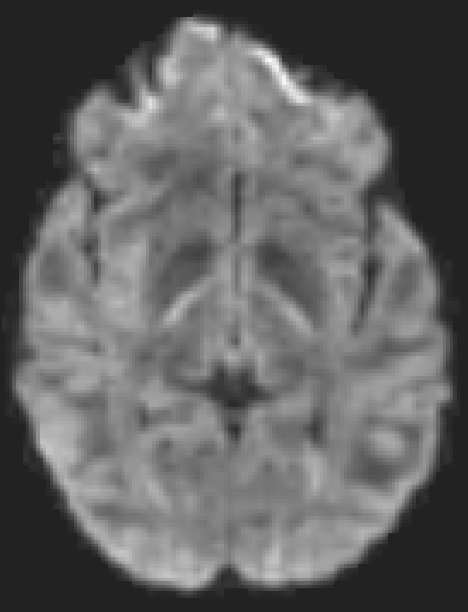}
        \label{fig:grad1}
    \end{subfigure}
    \hspace{1mm}
    \begin{subfigure}[b]{0.15\textwidth}
        \centering
        \includegraphics[width=\textwidth]{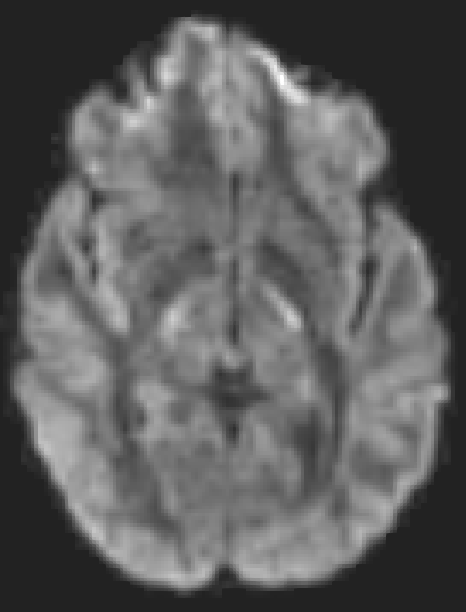}
        \label{fig:grad2}
    \end{subfigure}
    \hspace{1mm}
    \begin{subfigure}[b]{0.15\textwidth}
        \centering
        \includegraphics[width=\textwidth]{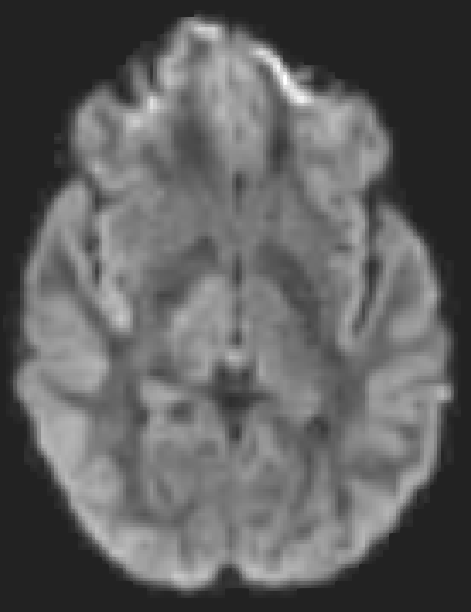}
        \label{fig:grad3}
    \end{subfigure}
    \hspace{1mm}
    \begin{subfigure}[b]{0.15\textwidth}
        \centering
        \includegraphics[width=\textwidth]{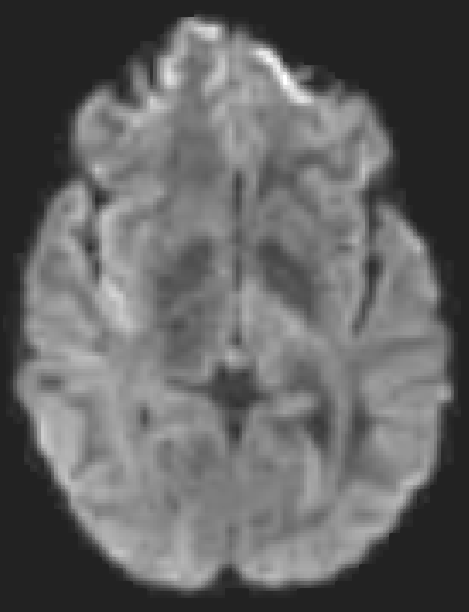}
        \label{fig:grad4}
    \end{subfigure}
    \hspace{1mm}
    \begin{subfigure}[b]{0.15\textwidth}
        \centering
        \includegraphics[width=\textwidth]{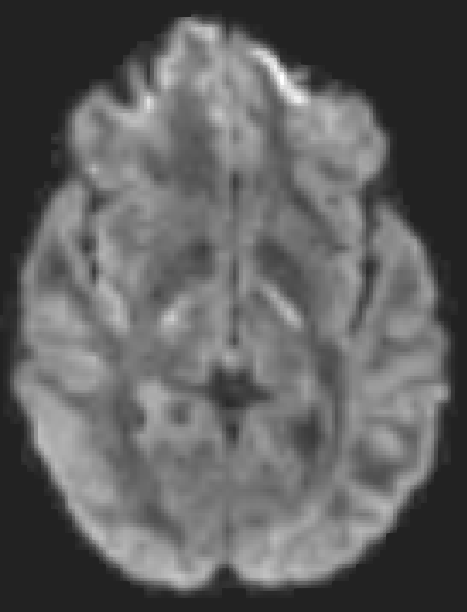}
        \label{fig:grad5}
    \end{subfigure}
    \hspace{1mm}
    \begin{subfigure}[b]{0.15\textwidth}
        \centering
        \includegraphics[width=\textwidth]{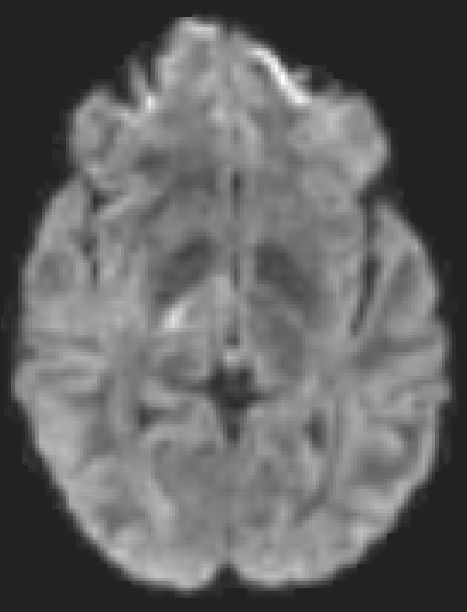}
        \label{fig:grad6}
    \end{subfigure}

    \caption{Axial slices from a volumetric \ac{dwi} dataset of a single subject where each scan was acquired at a different diffusion gradient direction. Data source: sub-1024 \ac{dwi} from TractoInferno dataset~\cite{tractinferno}}.
    \label{fig:mri_views}
\end{figure}

Streamline tractography aims to reconstruct white matter pathways by inferring plausible fiber trajectories from diffusion MRI data. We frame this problem as a sequential prediction task, where the likelihood of fiber orientations at a given point along a streamline is predicted based on the history of all previous \ac{dwi} measurements along that path.
Our dataset consists of \ac{dwi} scans and the corresponding tractography data of \(N\) subjects. For each subject, we have:
\begin{enumerate}[leftmargin=18pt]
    \item A 4D \ac{dwi} volume \(\boldsymbol{X}\in\mathbb{R}^{H\times W\times D \times G}\), where \(H\),\(W\), and \(D\) are spacial dimensions, and \(G\) corresponds to the number of magnetic field gradient directions applied during the \ac{dmri} scan.
     Axial views extracted from a volumetric \ac{dwi} dataset of a single subject, acquired at six (out of 65) different gradient directions are shown in Figure~\ref{fig:mri_views}. 
    \item A set of reference streamlines \(\mathcal{S} = \left\{\boldsymbol{s}^{(1)},\dots,\boldsymbol{s}^{(M)}\right\}\), representing a whole-brain tractography corresponding to \(\boldsymbol{X}\), where each streamline \(\boldsymbol{s}^{(m)} = ({s}^{m}_1,{s}^{m}_2,\dots,{s}^{m}_{N_m})\) is a sequence of 3D points in RAS (Right-Anterior-Superior) coordinates, commonly used in to standardize anatomical positions.
\end{enumerate}

We feed our model with sequences of \ac{dwi} values sampled along the coordinate path of a streamline. That is, given a streamline \(\boldsymbol{s} = ({s}_1,\dots,{s}_n)\), the input to the model is the sequence \(\left\{\boldsymbol{X}({\boldsymbol{s}_{1}}), \dots, \boldsymbol{X}({\boldsymbol{s}_{n}})\right\}\).

At each point along a streamline, the model is trained to predict an \ac{fodf}, represented as a discrete probability distribution over a fixed set of \( K + 1 \) classes. Here, \( K \) denotes a set of directions uniformly distributed on the unit sphere. Specifically, given a prefix trajectory \( (s_1, \dots, s_{i-1}) \), the output at point \( s_i \) is:
\begin{equation}
    \mathbb{P}(\boldsymbol{f}\mid \boldsymbol{X}(s_1),\dots, \boldsymbol{X}(s_i)),
\end{equation}
where \(\boldsymbol{f} = (f_1, \dots, f_K)\) is a discrete probability distribution over the direction classes defined by the spherical tessellation of \(K = 724\) directions, along with an additional class representing \ac{eof}. This conditional formulation reflects the core assumption of our model: the fiber orientation at a given point depends not only on the local microstructural context (captured by the \ac{dwi} signal), but also on the trajectory taken to reach that point.

\subsection{Model Architecture}
\label{arch}

\begin{figure}[ht]
    \centering
    \includegraphics[width=\linewidth]{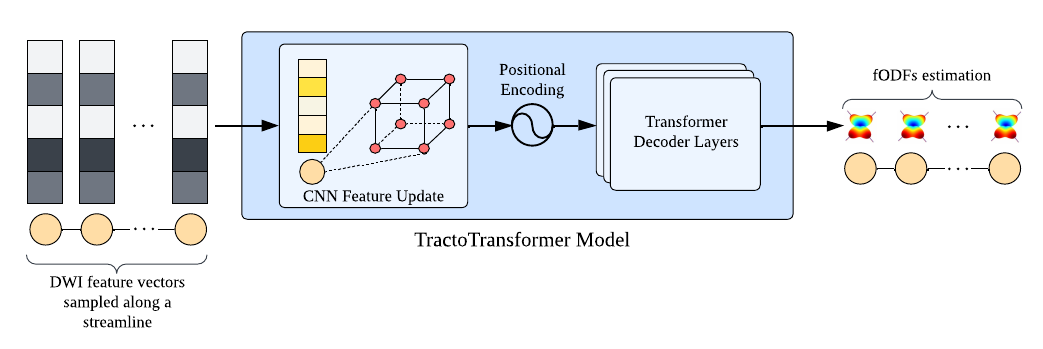}
    \caption{Overview of the TractoTransformer model framework. Streamlines are represented as sequences of \ac{dwi} feature vectors, where each vector is derived from the raw \ac{dmri} data of a specific voxel, sampled using spherical harmonics. The TractoTransformer model consists of a 3D-CNN layer, positional encoder, and Transformer layers. The CNN is applied to each streamline voxel (represented by a diffusion measurement vector) and its nearby spatial neighbors. The entire encoded streamline is fed into the transformer. The model outputs predicted \acp{fodf} at each voxel, which are used to guide subsequent tractography.}
    \label{fig:nn-diagram}
\end{figure}

The proposed TractoTransformer leverages the strengths of both Transformers and convolutional neural networks (CNNs) to predict conditional \acp{fodf} from \ac{dwi} data. Its architecture is illustrated in Figure~\ref{fig:nn-diagram}.

\textbf{3D Input Embedding.} 
To embed the input sequence for the Transformer, we first enhance each voxel representation along a streamline by incorporating local spatial context using a \ac{cnn3d}. For each point in the sequence, the \ac{cnn3d} processes a surrounding voxel cube to extract microstructural features from the local diffusion signal. This step improves the voxel-wise representation and expands the effective receptive field, providing the model with spatial context. To reduce computational cost, the \ac{cnn3d} is applied only to the batch of voxels corresponding to the current set of streamlines, rather than the entire brain volume.

The resulting spatially enhanced feature vectors serve as input tokens to the Transformer. To preserve the sequential order of streamline points, we apply standard sinusoidal positional encodings, as introduced by Vaswani et al. \cite{TRANSFORMERS}. This enables the model to account for trajectory history, ensuring that each orientation prediction is informed not only by local voxel features but also by the path taken to reach that point—an important consideration for anatomically plausible tractography.

\textbf{Decoder-Only Transformer.}
We use a standard decoder-only Transformer architecture to process sequences of streamline data. Each decoder block includes masked multi-head self-attention and a position-wise feed-forward network, both followed by residual connections and layer normalization. A causal attention mask enforces autoregressive prediction by preventing access to future positions, while a padding mask blocks attention to invalid inputs. This design allows the model to capture long-range dependencies and contextual patterns along the streamline. The final output is mapped to the target space via fully connected layers, followed by a softmax function that yields a probability distribution over possible directions and an end-of-fiber (EoF) class.
\subsection{Model Optimization and Loss Function}
\label{optim}
We use the reference streamlines provided in the dataset to construct labels for supervised learning, training the model to predict conditional \acp{fodf} during sequence processing. For each reference streamline, direction vectors are computed between consecutive points and normalized to unit vectors. Since the output classes correspond to directions on the unit sphere and possess a geometric structure with well-defined angular distances, it is appropriate to weigh classification errors accordingly~\cite{DEEPTRACT}.

To this end, we construct a soft label distribution by smoothing each unit direction over the sphere using a Gaussian kernel. Formally, given a unit direction \(\theta\) (i.e., the direction between two consecutive streamline points) and a set of unit directions \(\{\alpha_i\}_{i=1}^{K}\) defining the discrete class space, we compute the angular distance \(d_i\) between \(\theta\) and each \(\alpha_i\), and assign weights as:

\begin{equation}
    w_i = \exp\left(-\frac{d_i^2}{2\sigma^2}\right),
\end{equation}

where \(\sigma\) is the standard deviation of the Gaussian kernel. The resulting soft label is a normalized probability distribution over directions:

\begin{equation}
    y_{\text{smooth}}[i] = \frac{w_i}{\sum_{j=1}^{K} w_j}.
\end{equation}

\begin{figure}
  \begin{minipage}[c]{0.45\textwidth}
    \caption{
       Visualization of the smoothed label distribution on the unit sphere. The generated distribution decays as the distance on the unit sphere increases, providing a probabilistic framework for supervising the model's \ac{fodf} predictions.
    } \label{fig:smoothlabels}
  \end{minipage}
  \begin{minipage}[c]{0.45\textwidth}
    \includegraphics[width=5cm]{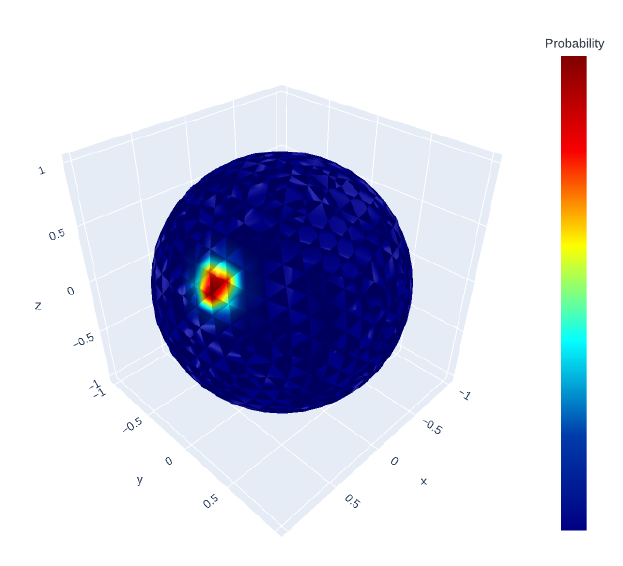}
  \end{minipage}
\end{figure}

This distribution decays smoothly with increasing angular distance on the unit sphere, as illustrated in Figure~\ref{fig:smoothlabels}, and is used to supervise the model’s predictions.

To train our TractoTransformer model, we employ the \ac{kld} loss to measure the divergence between the predicted discrete distribution and the corresponding smooth target label. Given a point \(s_i\), a target label distribution \( y_{\text{smooth}} \) associated with \( s_i \), and a model prediction \( y_{\text{pred}} = \mathbb{P}(\boldsymbol{f} \mid \boldsymbol{X}(s_1), \dots, \boldsymbol{X}(s_i)) \), the \ac{kld} is formally defined as:
\begin{equation}
    \mathcal{L}_{\text{KL}}(y_{\text{smooth}}, y_{\text{pred}}) = \sum_{j=1}^{K}y_{\text{smooth}}[j] \log\left(\frac{y_{smooth}[j]}{\mathbb{P}\left({f_j\mid\boldsymbol{X}(s_1),\dots,\boldsymbol{X}(s_i)}\right)}\right),
\end{equation}
The mean loss is computed at each prediction step along the streamline. The \ac{kld} loss quantifies the information loss incurred when \( y_{\text{pred}} \) is used to approximate \( y_{\text{true}} \), making it well-suited for evaluating the accuracy of probabilistic predictions against the ground truth distribution of fiber orientations. This loss function is particularly appropriate in scenarios where both the predicted outputs and the target labels are probability distributions, as it encourages the model to produce outputs that closely align with the empirical data.

\subsection{Streamline Tractography Inference}
\label{inference}
Once the model is trained, tractography is initiated by sampling random seed points from the provided white matter mask, each defining the starting location of a fiber trajectory. Tractography proceeds iteratively: at each step, the model receives the current point along with the accumulated tracking history and predicts a conditional \ac{fodf}, auto-regressively conditioned on previously generated points in the streamline. This design ensures that each orientation prediction incorporates both local features and the full trajectory context, capturing the sequential dependencies inherent in white matter pathways.

The tracking direction is selected as the one with the highest probability in the predicted \ac{fodf}, resulting in a deterministic propagation scheme. However, unlike classical deterministic methods, our predictions are context-aware—conditioned on the entire streamline history—enabling robust direction selection even in anatomically challenging regions such as fiber crossings or areas of high uncertainty.

After selecting a direction, the streamline is advanced by a fixed step in RAS space, the new point is appended to the trajectory, and the process is repeated until a stopping criterion is met:
\begin{enumerate}[leftmargin=18pt]
    \item The class chosen from the prediction of the model is \ac{eof} class.
    \item The next step is outside of the bounds of the MRI image.
    \item The next step is outside of the white matter mask.
    \item The angle between two consecutive steps exceeds a predefined threshold.
    \item The \ac{fa} values in the next step are less than a predefined threshold.
\end{enumerate}
The collection of generated trajectories constitutes a set of approximate streamlines which together form the final tractogram. This process is detailed in the pseudocode provided in Algorithm \ref{alg:tractography}.

\begin{algorithm}[ht]
\caption{Streamline Tractography Algorithm}
\label{alg:tractography}
\begin{algorithmic}[1]
\Require Trained model, white matter mask, seed points, stopping criteria
\Ensure Tractogram of streamlines
\For{each seed point}
    \State Initialize streamline with seed point
    \While{stopping criteria are not met}
        \State Feed the current streamline into the model
        \State Get conditional fODFs from the model
        \State Select direction as \texttt{argmax} of conditional fODF
        \State Compute next point by stepping in the selected direction
        \If{next point satisfies stopping criteria}
            \State Terminate streamline
        \Else
            \State Add next point to the streamline
        \EndIf
    \EndWhile
    \State Store the completed streamline in tractogram
\EndFor
\end{algorithmic}
\end{algorithm}


\section{Experiments}
\label{experimental}
\subsection{Datasets}
\label{dataset}

For this study, we used two publicly available tractography datasets. The first is the ISMRM 2015 Tractography Challenge phantom dataset~\cite{ISMRM}, which has been one of the most widely used benchmarks in the field over the past decade. It contains a high-quality 4D \ac{dwi} volume with dimensions \(90 \times 108 \times 90 \times 100\) after resampling, along with a comprehensive set of 270{,}000 ground truth white matter streamlines. 

The second dataset is TractoInferno~\cite{tractinferno}, the largest open-source, multi-site tractography dataset to date, comprising diffusion data from 286 subjects. The dataset is partitioned into 198 subjects for training, 60 for validation, and 28 for testing. Each subject includes a 4D diffusion-weighted imaging (DWI) volume with higher spatial resolution than the ISMRM dataset. Although the exact dimensions vary between subjects (for example, the test subject whose tractography is shown in Figure~\ref{fig:unified_tractography_comparison} has a spatial volume of \(141 \times 184 \times 120\)), the number of gradient directions also varies, ranging from 22 to 132, and is resampled to 100 directions for consistency.  In addition, the dataset provides rich ground-truth tractography, averaging over 1 million streamlines per subject. These streamlines are generally shorter than those in the ISMRM dataset and, after resampling to a constant step size of \(3\,\mathrm{mm}\), can contain up to 100 3D points.

\subsection{Preprocessing.}
\label{preprocess}
To ensure consistency across subjects and reduce variability arising from acquisition protocols, we apply several preprocessing steps. First, we represent the \ac{dwi} signal using spherical harmonic coefficients sampled over a fixed set of gradient directions. This step addresses inter-subject variability in gradient schemes and provides a standardized input format for the model. Next, we resample the reference streamlines to maintain a consistent step size between consecutive points in the right-anterior-superior (RAS) space, ensuring uniform spatial resolution across all samples and supporting reliable modeling and downstream analysis. Finally, we augment the dataset by reversing streamline orientations, increasing data diversity, and enabling the model to learn more robust features.
\subsection{Implementation Details}
The model configurations are as follows. For the \ac{cnn3d} component, we use a single 3D convolutional layer with a kernel size of \(3 \times 3 \times 3\). The \ac{dwi} data was resampled to 100 fixed gradient directions. To mitigate overfitting, dropout with a probability of \(p = 0.1\) was applied to all layers. For label smoothing, we employed a Gaussian kernel with a standard deviation of \(\sigma = 0.1\). Target labels were represented as discrete probability distributions over 725 classes (\(K = 724\)), corresponding to an angular resolution of approximately \(3.5^\circ\).

The Transformer-based network consists of 8 decoder layers, each with 10 attention heads. Every decoder block includes a \ac{ffn} with a hidden dimension of 512. The final Transformer output is passed through an additional \ac{ffn}, projecting it to a 725-dimensional vector representing 724 candidate directions on the unit sphere and one end-of-fiber (\ac{eof}) class used to signal streamline termination.

All models were trained using the Adam optimizer~\cite{ADAM} with an initial learning rate of 0.005. Learning rate decay was applied by multiplying the rate by 0.7 if the accuracy did not improve by at least 0.3 over two consecutive epochs. Training was conducted for 30 epochs with a batch size of 20, using up to four NVIDIA V100 GPUs with 32GB of memory. For inference, we used an angular threshold of 70 degrees and an \ac{fa} threshold of 0.05.

\subsection{Quantitative Evaluation on ISMRM Dataset}
To evaluate our model, we trained it on the ISMRM dataset using an 80/20 split of reference streamlines for training and validation. Training took 12 hours. Whole-brain tractography was then performed by seeding from random points within the white matter mask. The resulting tractograms were compared to ground truth and a state-of-the-art method using the Tractometer tool\cite{ISMRM}, a standardized evaluation framework.

We report four key metrics: \textbf{\ac{vc}} (valid connection rate), \textbf{\ac{ol}} (overlap with ground truth), \textbf{\ac{or}} (overreach beyond anatomical boundaries), and the \textbf{F1 score}, which balances precision and recall.
\begin{table}[h]
  \caption{Tractometer evaluation results for various state-of-the-art methods. The best-performing values for each metric are highlighted in bold, and the second-best are underlined. Our method achieves the highest scores in \ac{vc}, \ac{ol}, and F1 among all evaluated approaches.}
  \vspace{4pt}
  \label{tab:tractometer_results}
  \centering
  \begin{tabular}{lllll}
    \toprule
    \textbf{Model} & \textbf{VC (\%)\(\uparrow\)} & \textbf{OL (\%)\(\uparrow\)} & \textbf{OR (\%)\(\downarrow\)} & \textbf{F1 (\%)\(\uparrow\)} \\
    \midrule
    \textbf{ISMRM Mean} & 54 & 31 & 23 & 44 \\
    \textbf{RF \cite{RF, NEHER2017}} & 67 & \underline{75} & 31 & - \\
    \textbf{LearnToTrack \cite{LearnToTrack}} & 42 & 64 & 35 & 64 \\
    \textbf{DeepTract \cite{DEEPTRACT}} & \underline{71} & 69 & \underline{23} & \underline{70} \\
    \textbf{Entrack \cite{Entrack}} & 65 & 60 & 36 & 58 \\
    \textbf{Track-to-learn \cite{TrackToLearn}} & 68 & 62 & - & - \\
    \textbf{CTtrack \cite{CTtrack}} & 57 & 50 & \textbf{16} & 60 \\
    \textbf{TractoTransformer} & \textbf{82} & \textbf{84} & 31 & \textbf{75} \\
    \bottomrule
  \end{tabular}
\end{table}
Table~\ref{tab:tractometer_results} shows that TractoTransformer outperforms state-of-the-art tractography methods. It achieves a \ac{vc} of 82\% and an overlap of 84\%, reflecting accurate reconstruction of valid white matter connections with close alignment to the ground truth, enabled by its high angular resolution. Despite an overreach of 31\%, it attains the highest F1 score (75\%), demonstrating strong precision in delineating white matter pathways.
\subsection{Ablation Study}
\label{ablation}
To evaluate the contribution of each component, we conducted an ablation study by individually removing the \ac{cnn3d} module, reverse streamline augmentation, and label smoothing. Results in Table~\ref{tab:ablation} highlight the performance impact of each element.
\begin{table}[h]
  \caption{Ablation study of the TractoTransformer framework using the Tractometer tool.}
  \label{tab:ablation}
  \centering
  \vspace{4pt}
  \begin{tabular}{lllll}
    \toprule
    \textbf{Model} & \textbf{VC (\%)\(\uparrow\)} & \textbf{OL (\%)\(\uparrow\)} & \textbf{OR (\%)\(\downarrow\)} & \textbf{F1 (\%)\(\uparrow\)} \\
    \midrule
    \textbf{TractoTransformer} 
        & \textbf{81.51} & \textbf{83.72} & \textbf{30.83} & 74.78 \\
    \textbf{~~~-\ac{cnn3d}} 
        & 69.86{\tiny (-11.65)} & 80.70{\tiny (-3.02)} & 32.84{\tiny (+2.01)} & 72.66{\tiny (-2.12)} \\
    \textbf{~~~-Reverse~Streamlines} 
        & 79.81{\tiny (-1.70)} & 82.94{\tiny (-0.78)} & 33.08{\tiny (+2.25)} & 73.76{\tiny (-1.02)} \\
    \textbf{~~~-Smooth~Labels} 
        & 79.77{\tiny (-1.74)} & 82.81{\tiny (-0.91)} & 30.86{\tiny (+0.03)} & \textbf{74.85}{\tiny (+0.07)} \\
    
    \bottomrule
  \end{tabular}
\end{table}
The largest performance drop occurs when excluding the \ac{cnn3d} module, reducing \ac{vc} by \(11.65\%\) and modestly lowering the F1 score, underscoring the importance of local spatial context for accurate trajectory estimation. Removing reverse streamline augmentation yields a smaller decline, suggesting directional diversity aids regularization but is less critical. Omitting smooth labels has minimal impact on F1, with slight decreases in \ac{vc} and \ac{ol}. Overall, while all components contribute, the \ac{cnn3d} is key to anatomically plausible reconstructions.
\subsection{Qualitative in-vivo Tractography Results}
\label{invivo}
\begin{figure*}[ht]
    \centering
    \begin{subfigure}[t]{0.2\textwidth}
        \centering
        \includegraphics[width=\textwidth]{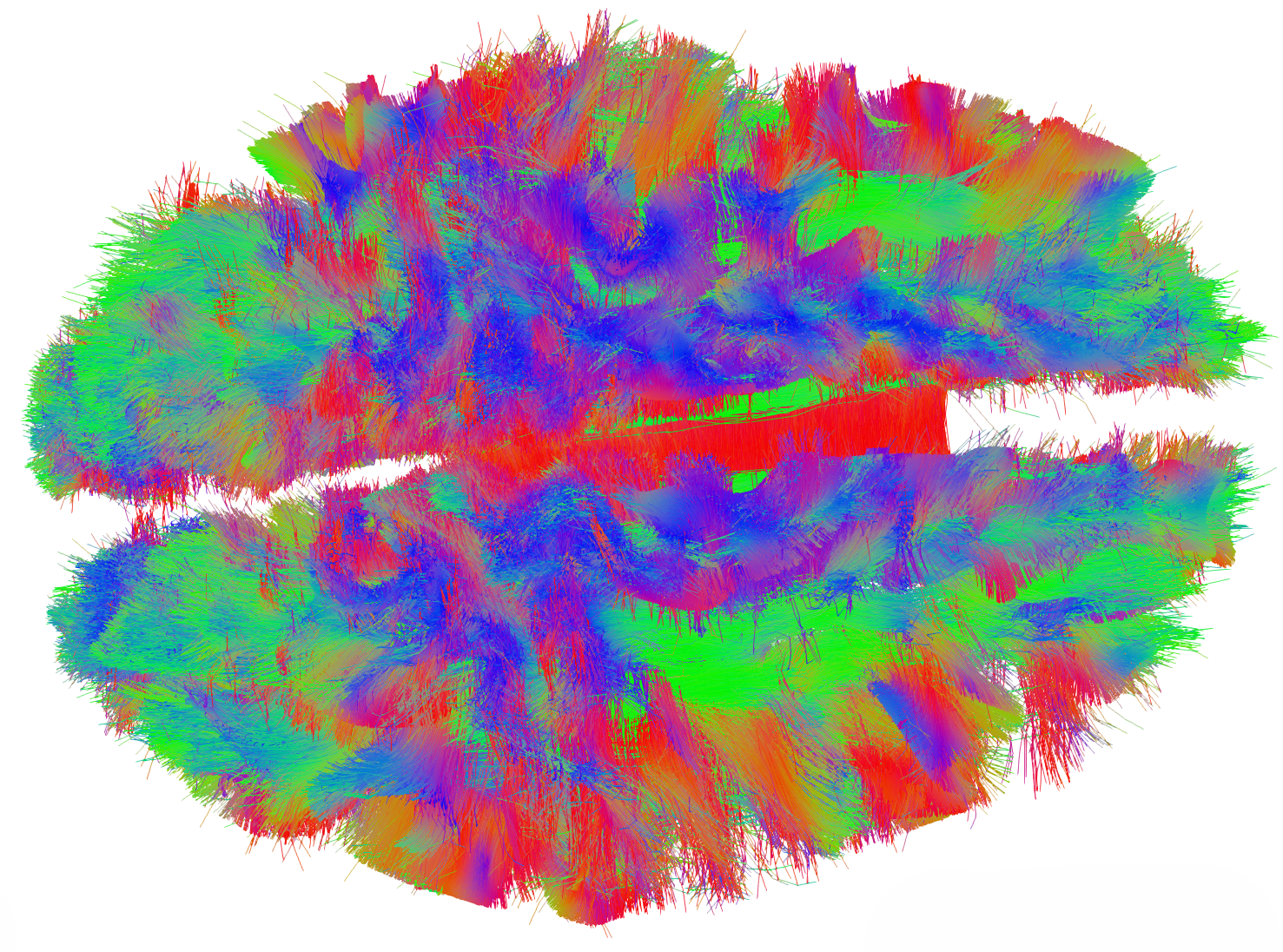}
        \caption*{GT \\ Axial view}
    \end{subfigure}
    \begin{subfigure}[t]{0.2\textwidth}
        \centering
        \includegraphics[width=\textwidth]{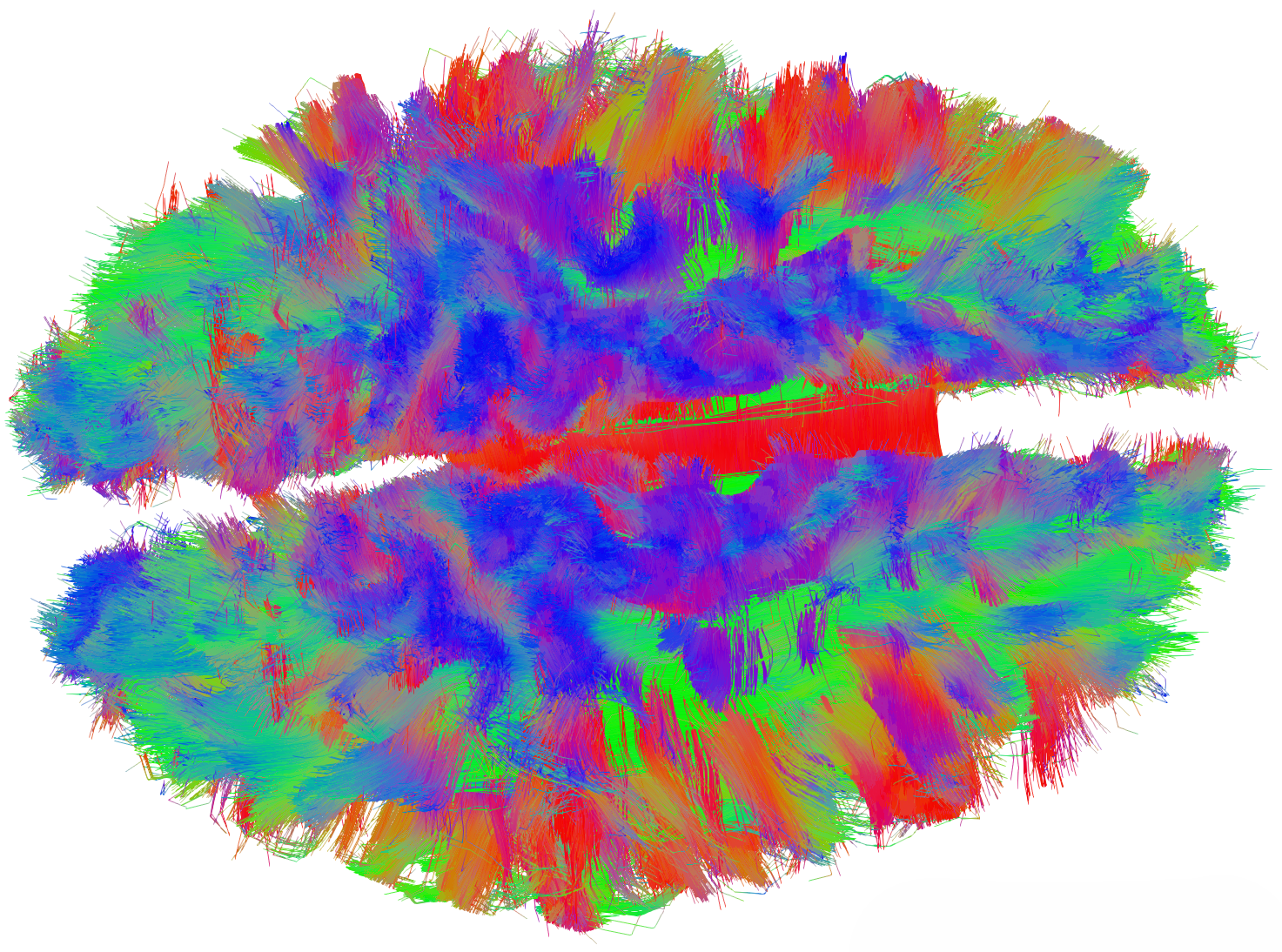}
        \caption*{Ours \\ Axial view}
    \end{subfigure}
    \hspace{0.02\textwidth}
    \begin{subfigure}[t]{0.2\textwidth}
        \centering
        \includegraphics[width=\textwidth]{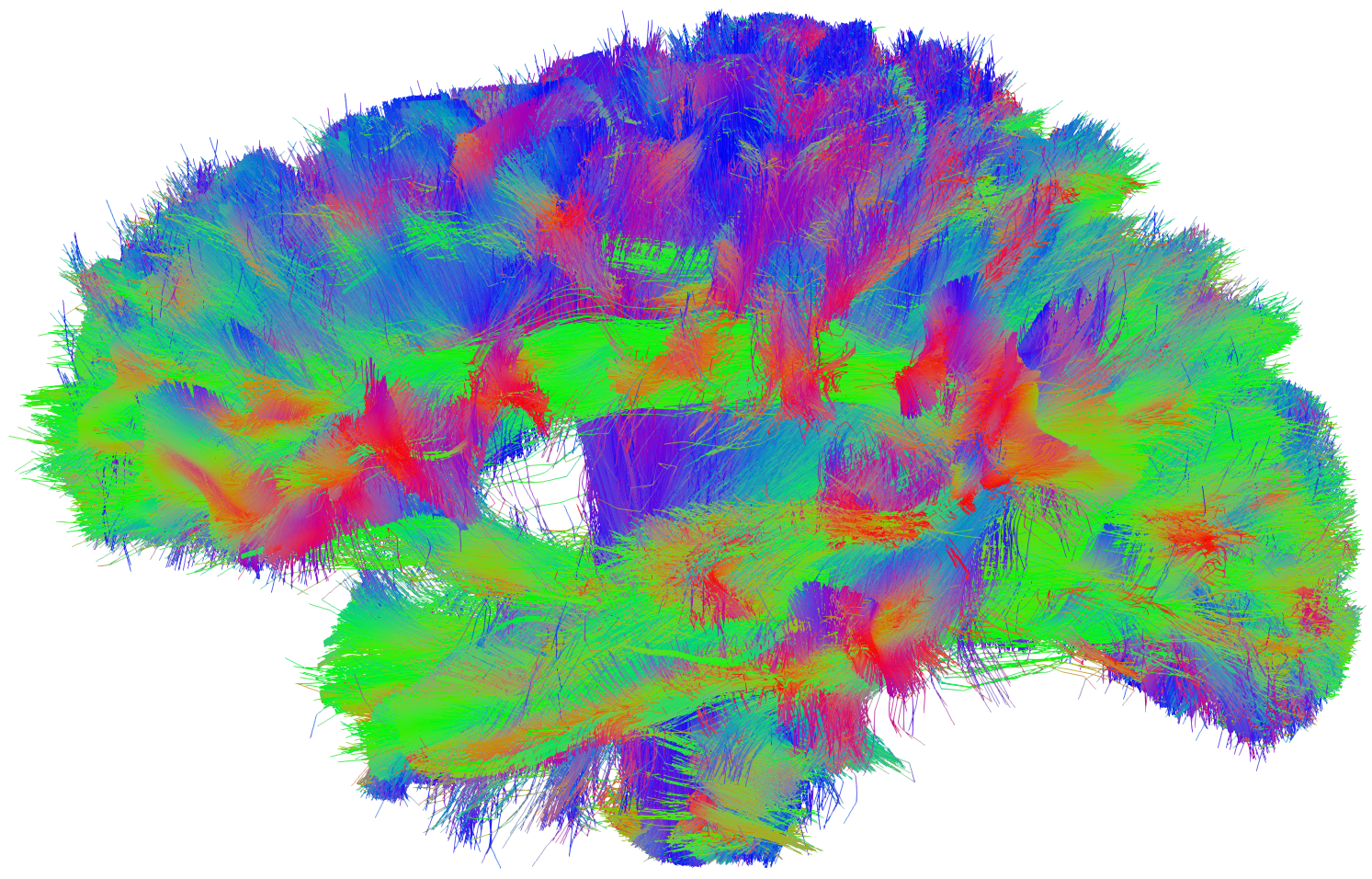}
        \caption*{GT \\ Sagittal view}
    \end{subfigure}
    \begin{subfigure}[t]{0.2\textwidth}
        \centering
        \includegraphics[width=\textwidth]{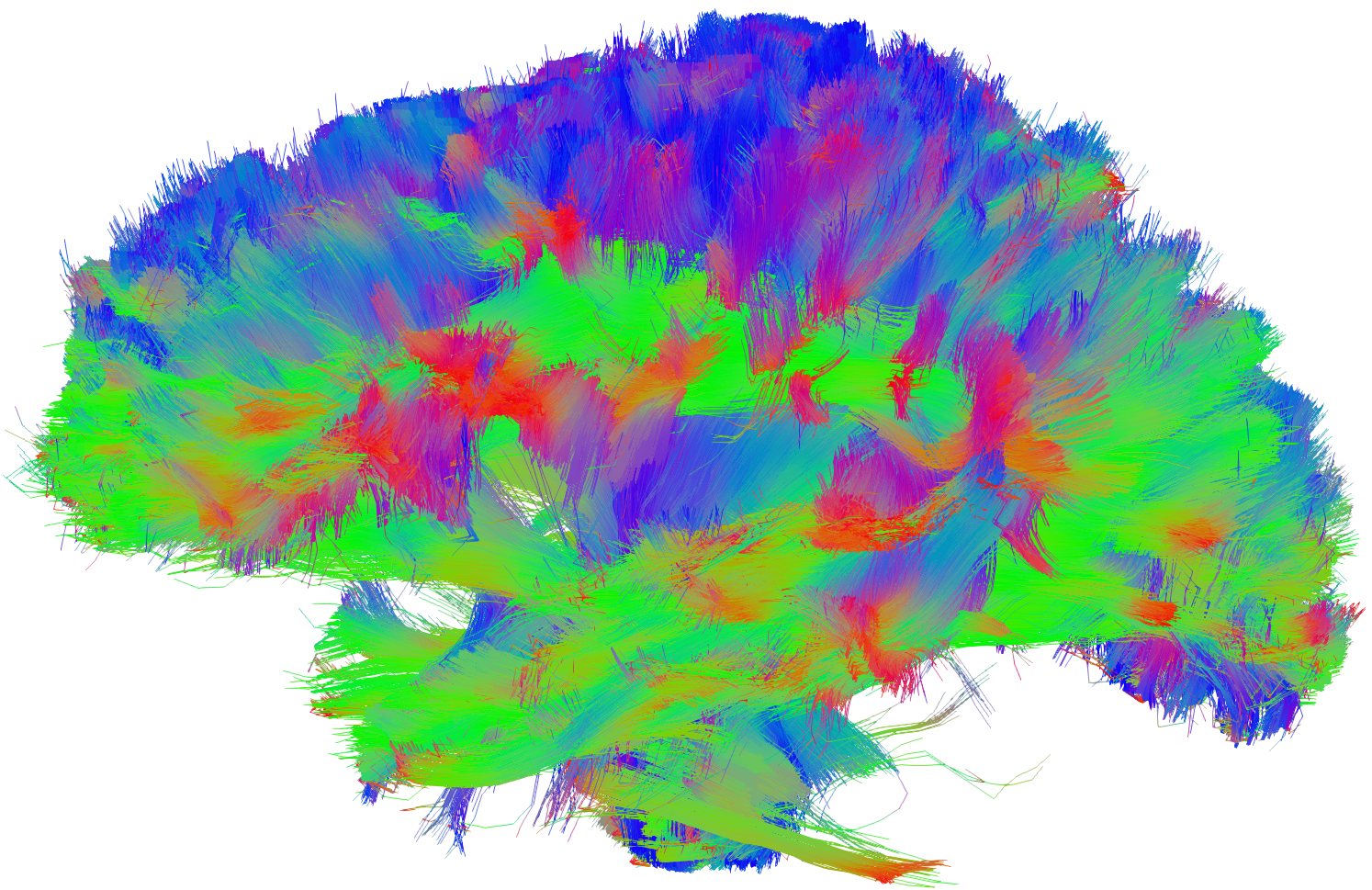}
        \caption*{Ours \\ Sagittal view}
    \end{subfigure}
    \hfill
    \begin{subfigure}[t]{0.15\textwidth}
        \centering
        \parbox[c][\textwidth][t]{\textwidth}{%
            \centering
            \textbf{Whole \\ Brain}
        }
    \end{subfigure}

    \vspace{0.7\baselineskip}

    \begin{subfigure}[t]{0.2\textwidth}
        \centering
        \includegraphics[width=\textwidth]{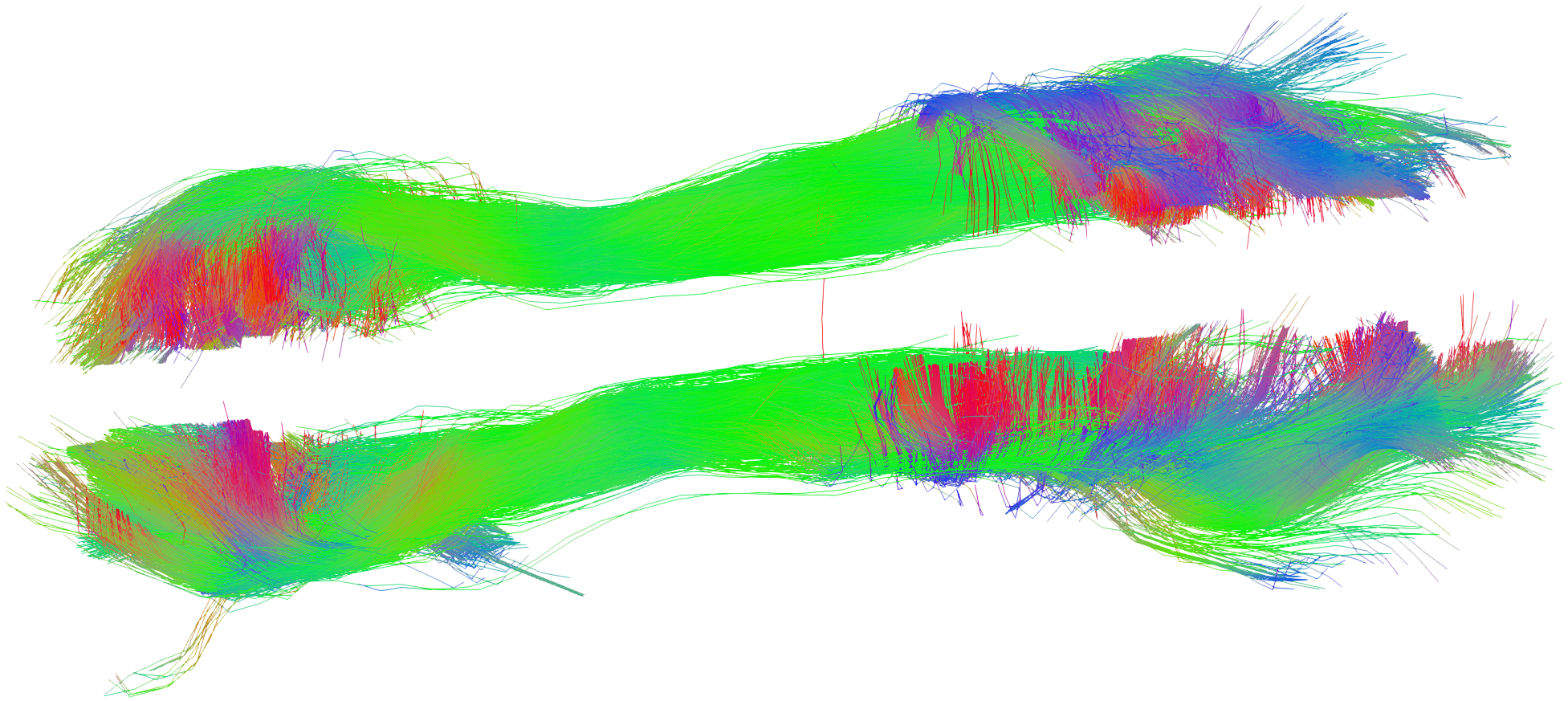}
        \caption*{GT \\ Axial view}
    \end{subfigure}
    \begin{subfigure}[t]{0.2\textwidth}
        \centering
        \includegraphics[width=\textwidth]{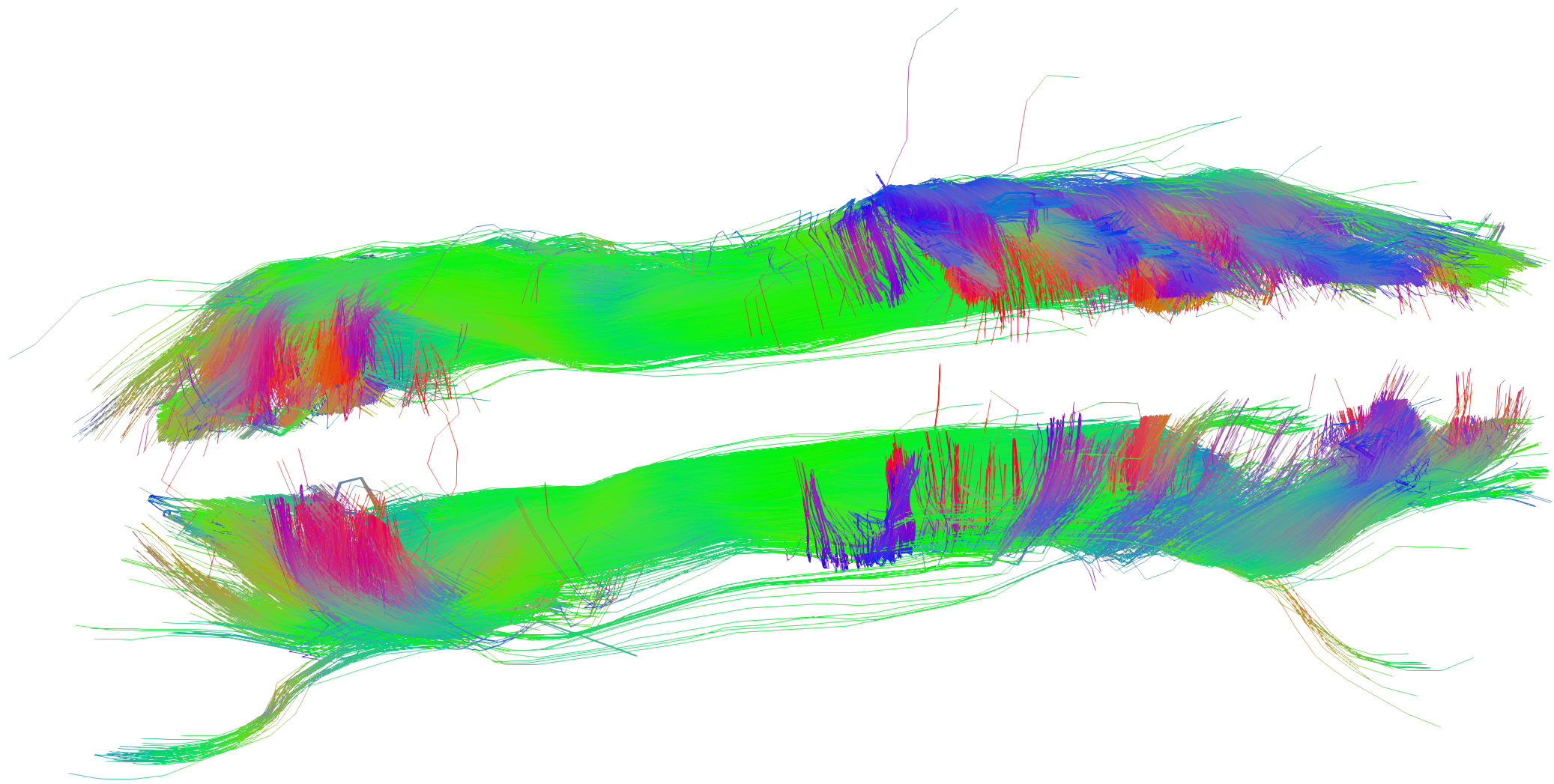}
        \caption*{Ours \\ Axial view}
    \end{subfigure}
    \hspace{0.02\textwidth}
    \begin{subfigure}[t]{0.2\textwidth}
        \centering
        \includegraphics[width=\textwidth]{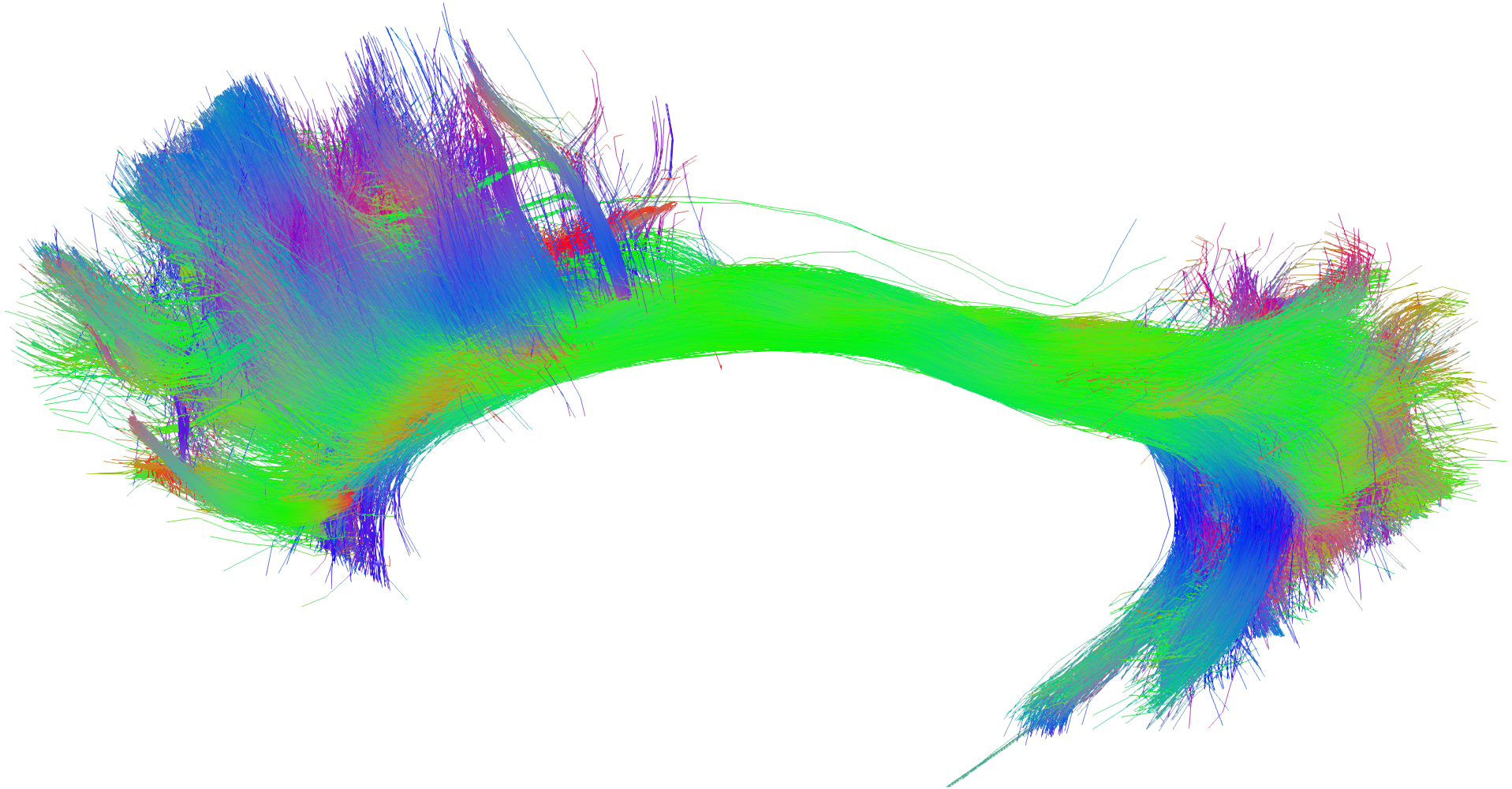}
        \caption*{GT \\ Sagittal view}
    \end{subfigure}
    \begin{subfigure}[t]{0.2\textwidth}
        \centering
        \includegraphics[width=\textwidth]{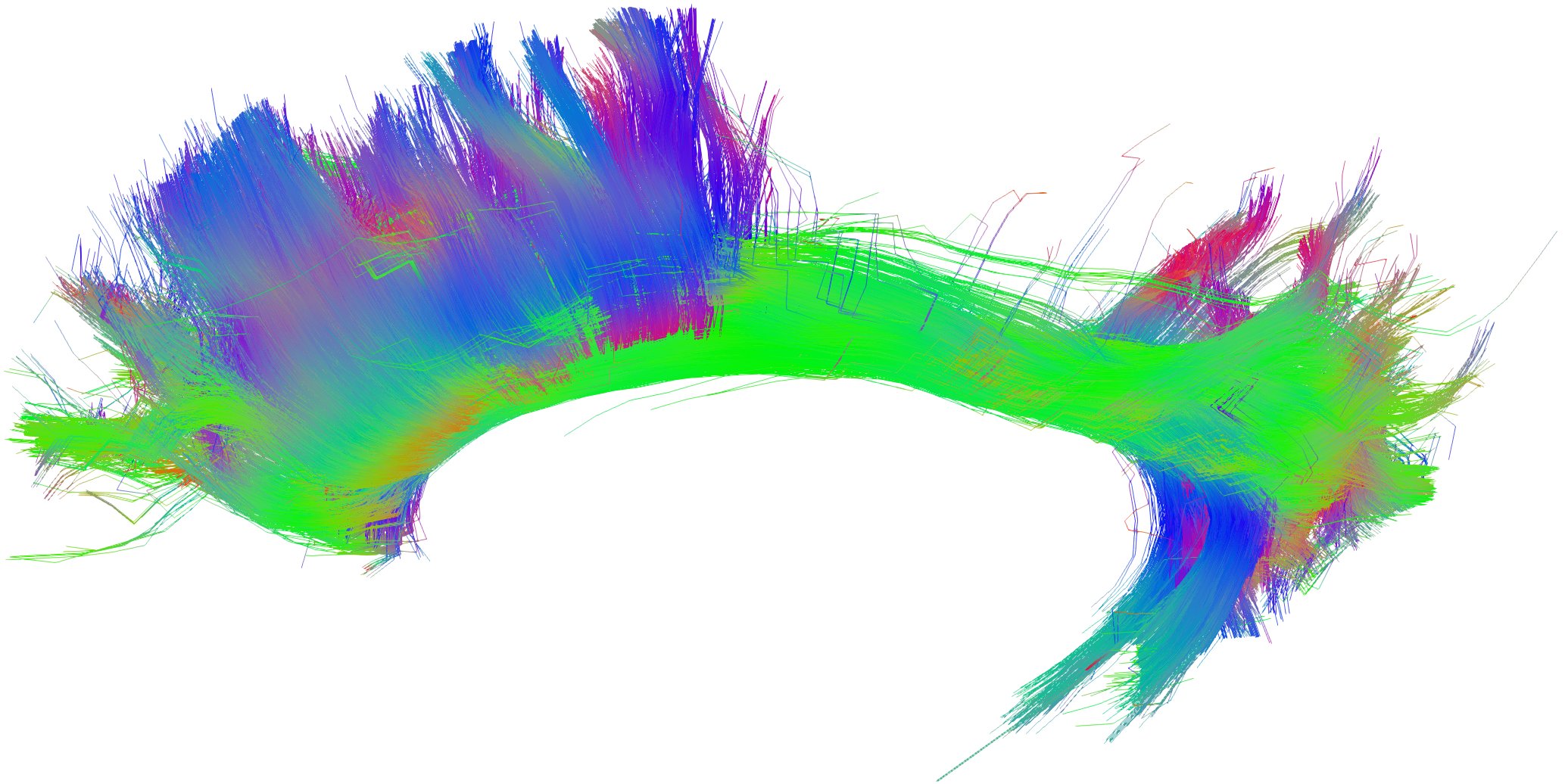}
        \caption*{Ours \\ Sagittal view}
    \end{subfigure}
    \hfill
    \begin{subfigure}[t]{0.15\textwidth}
        \centering
        \parbox[c][\textwidth][c]{\textwidth}{%
            \centering
            \vspace*{-1cm}
            \textbf{Cingulum}
        }
    \end{subfigure}

    \vspace{0.7\baselineskip}

    \begin{subfigure}[t]{0.19\textwidth}
        \centering
        \includegraphics[width=\textwidth]{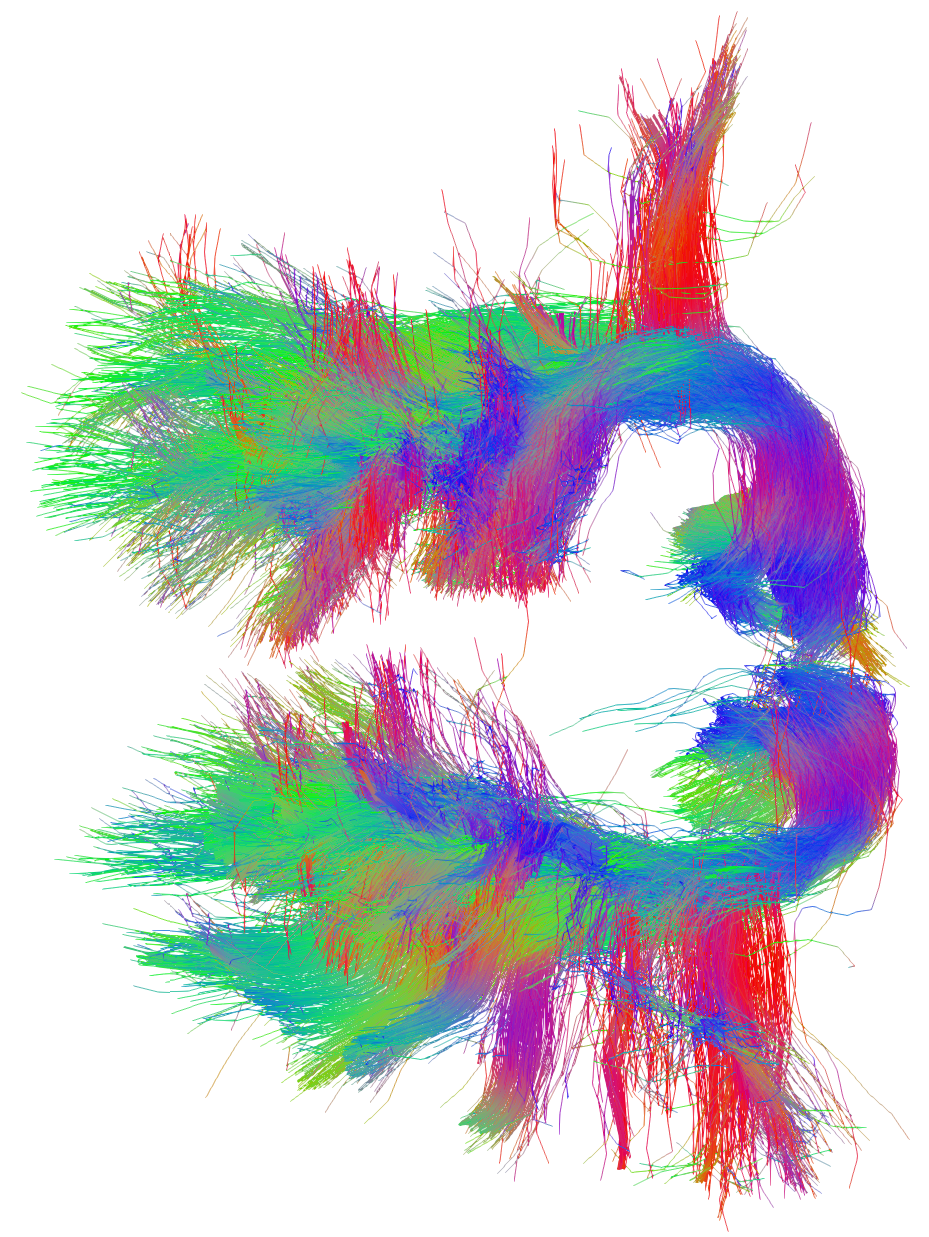}
        \caption*{GT \\ Axial view}
    \end{subfigure}
    \begin{subfigure}[t]{0.19\textwidth}
        \centering
        \includegraphics[width=0.95\textwidth]{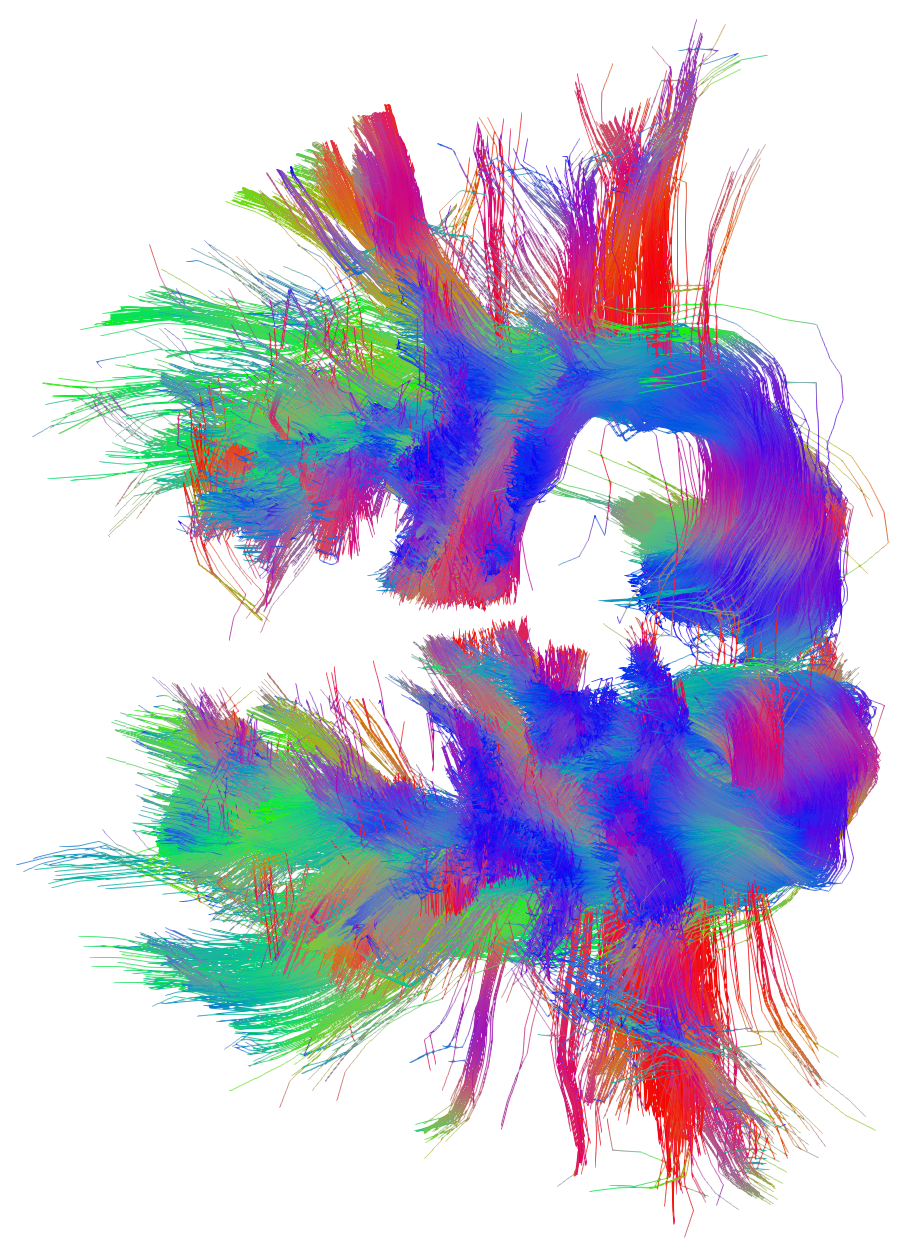}
        \caption*{Ours \\ Axial view}
    \end{subfigure}
    \hspace{0.02\textwidth}
    \begin{subfigure}[t]{0.21\textwidth}
        \centering
        \includegraphics[width=\textwidth]{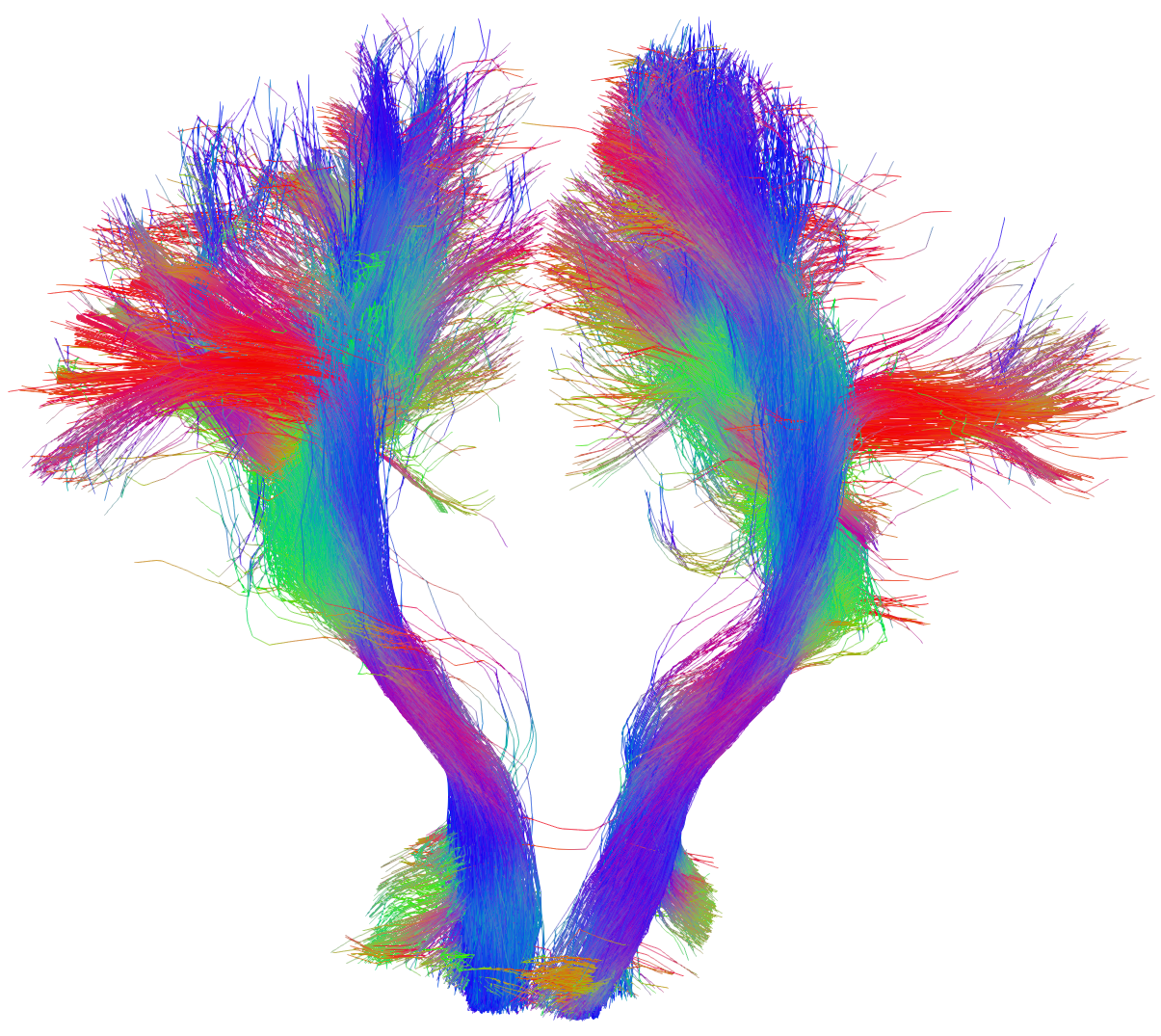}
        \caption*{GT \\ Coronal view}
    \end{subfigure}
    \begin{subfigure}[t]{0.21\textwidth}
        \centering
        \includegraphics[width=\textwidth]{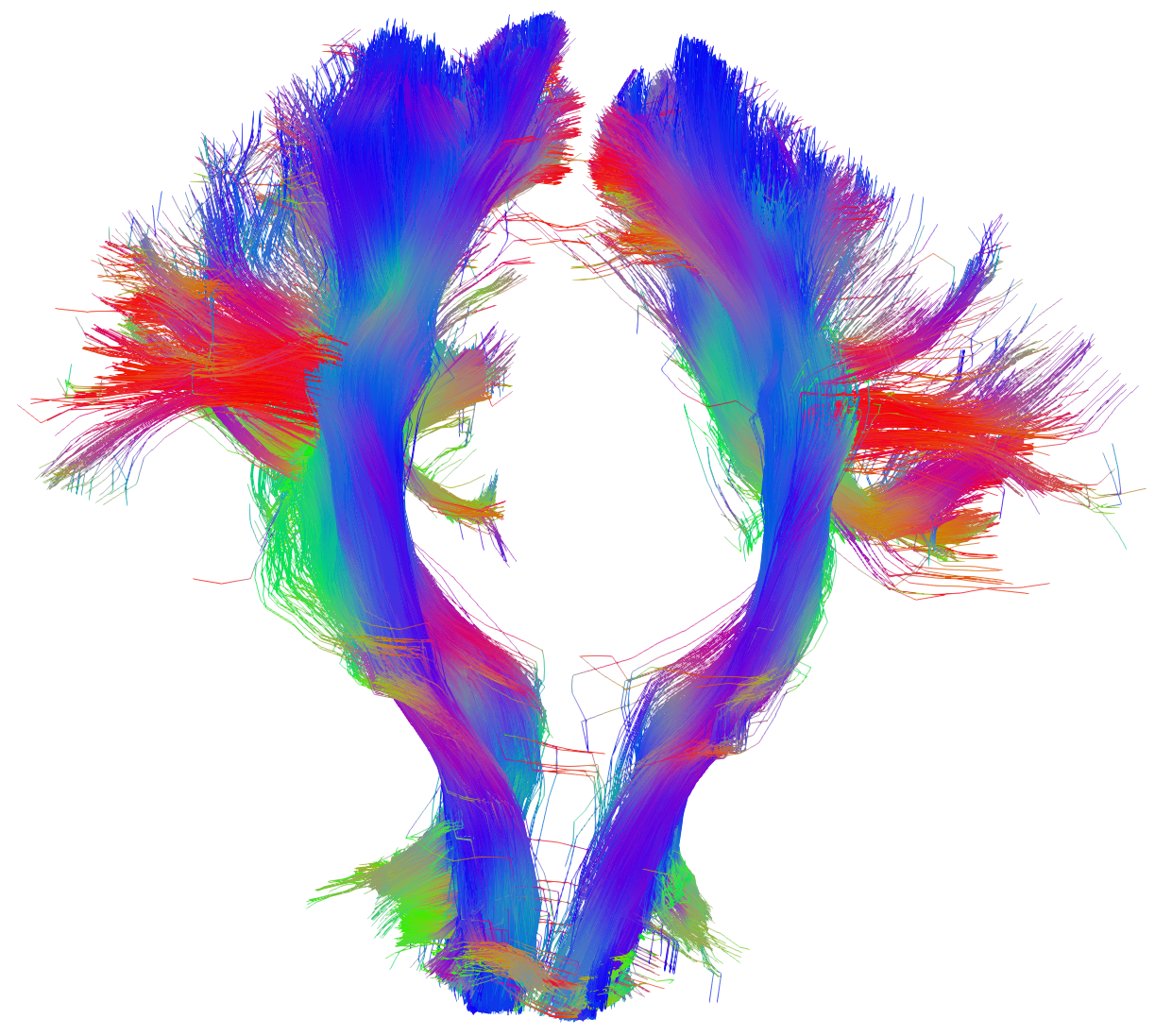}
        \caption*{Ours \\ Coronal view}
    \end{subfigure}
    \hfill
    \begin{subfigure}[t]{0.15\textwidth}
        \centering
        \parbox[c][\textwidth][t]{\textwidth}{%
            \centering
            \vspace*{-1cm}
            \textbf{Parieto-Occipital Pontine}
        }
    \end{subfigure}

    \vspace{0.7\baselineskip}

    \begin{subfigure}[t]{0.2\textwidth}
        \centering
        \includegraphics[width=\textwidth]{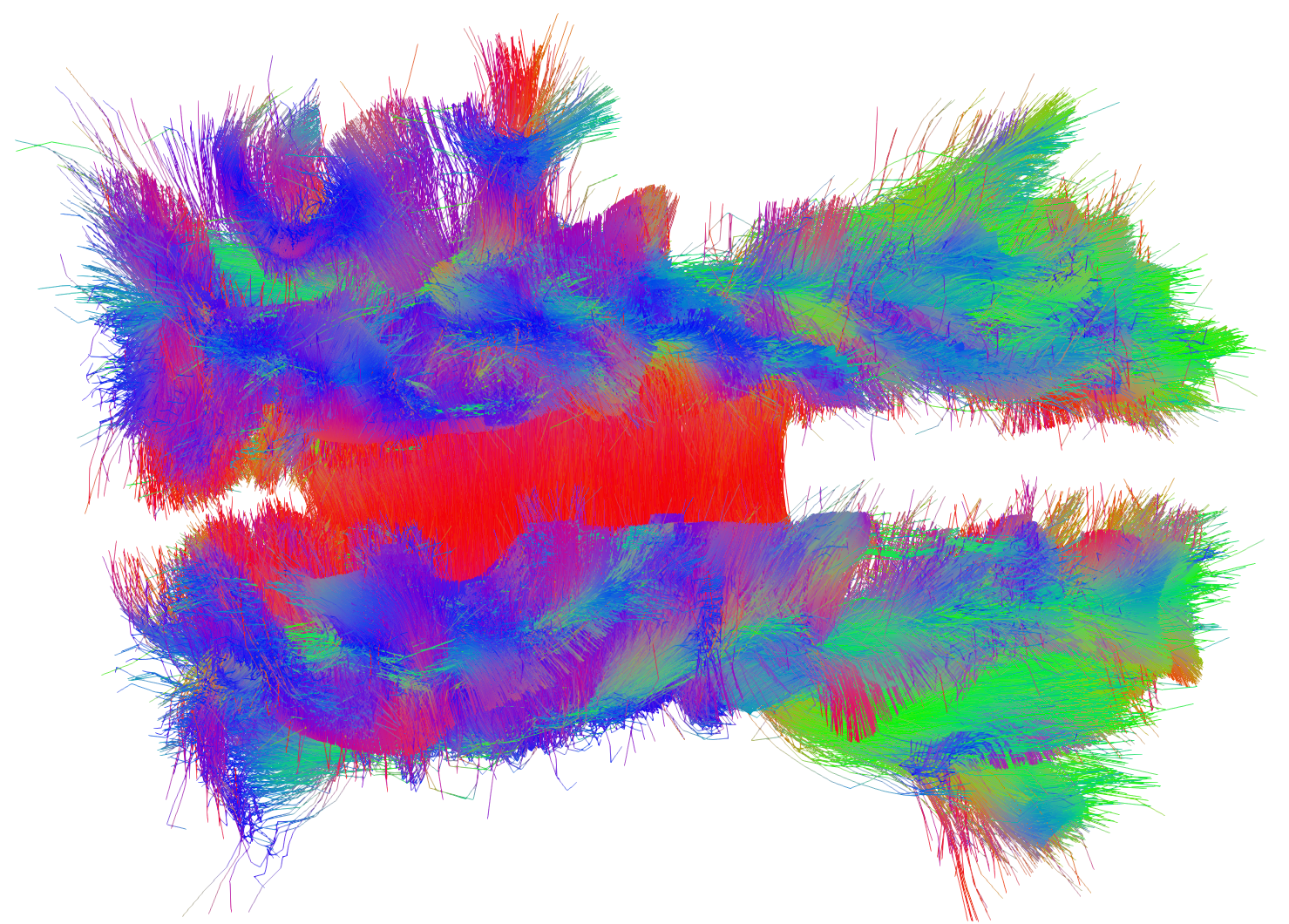}
        \caption*{GT \\ Axial view}
    \end{subfigure}
    \begin{subfigure}[t]{0.2\textwidth}
        \centering
        \includegraphics[width=0.87\textwidth]{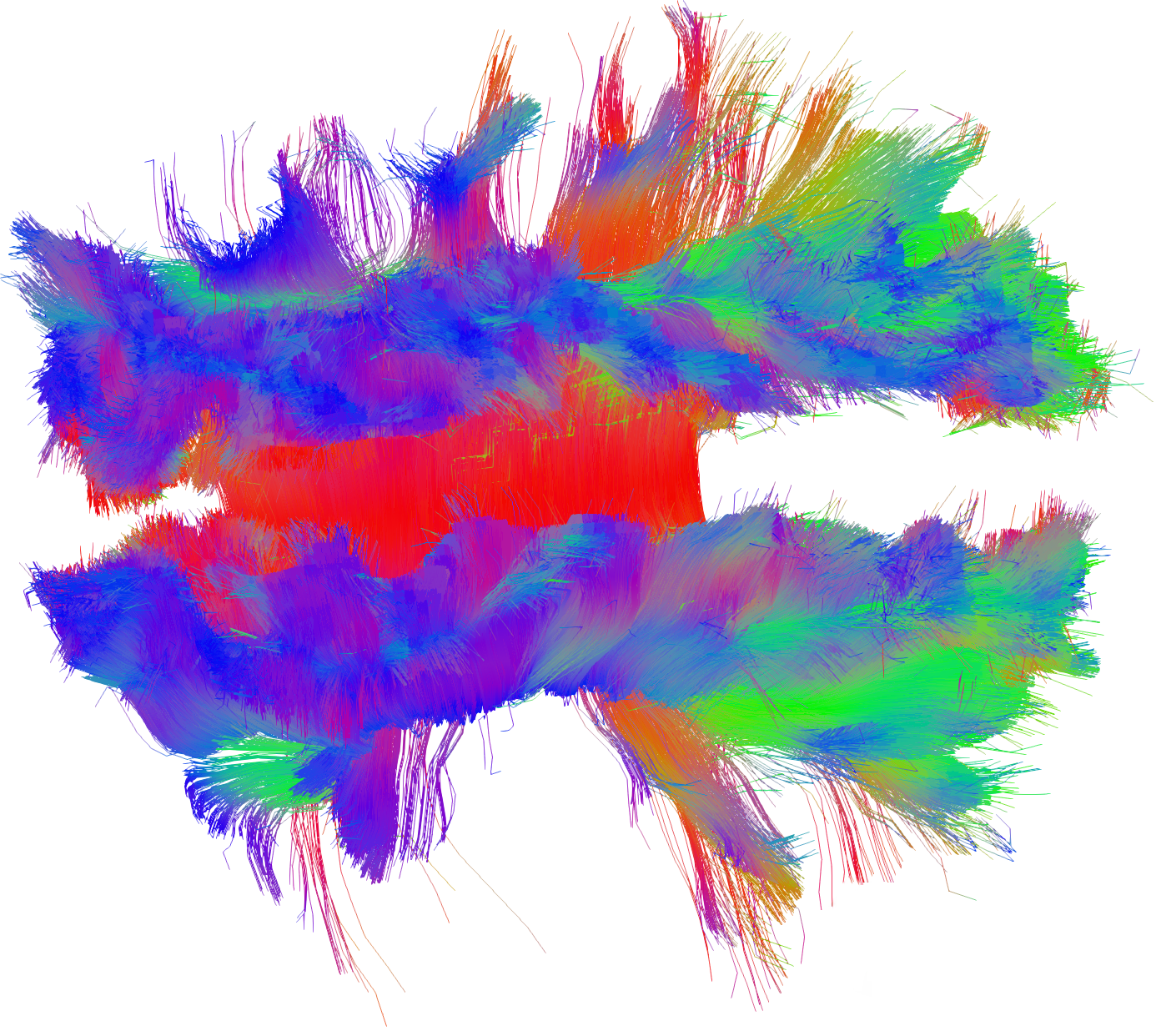}
        \caption*{Ours \\ Axial view}
    \end{subfigure}
    \hspace{0.02\textwidth}
    \begin{subfigure}[t]{0.2\textwidth}
        \centering
        \includegraphics[width=0.83\textwidth]{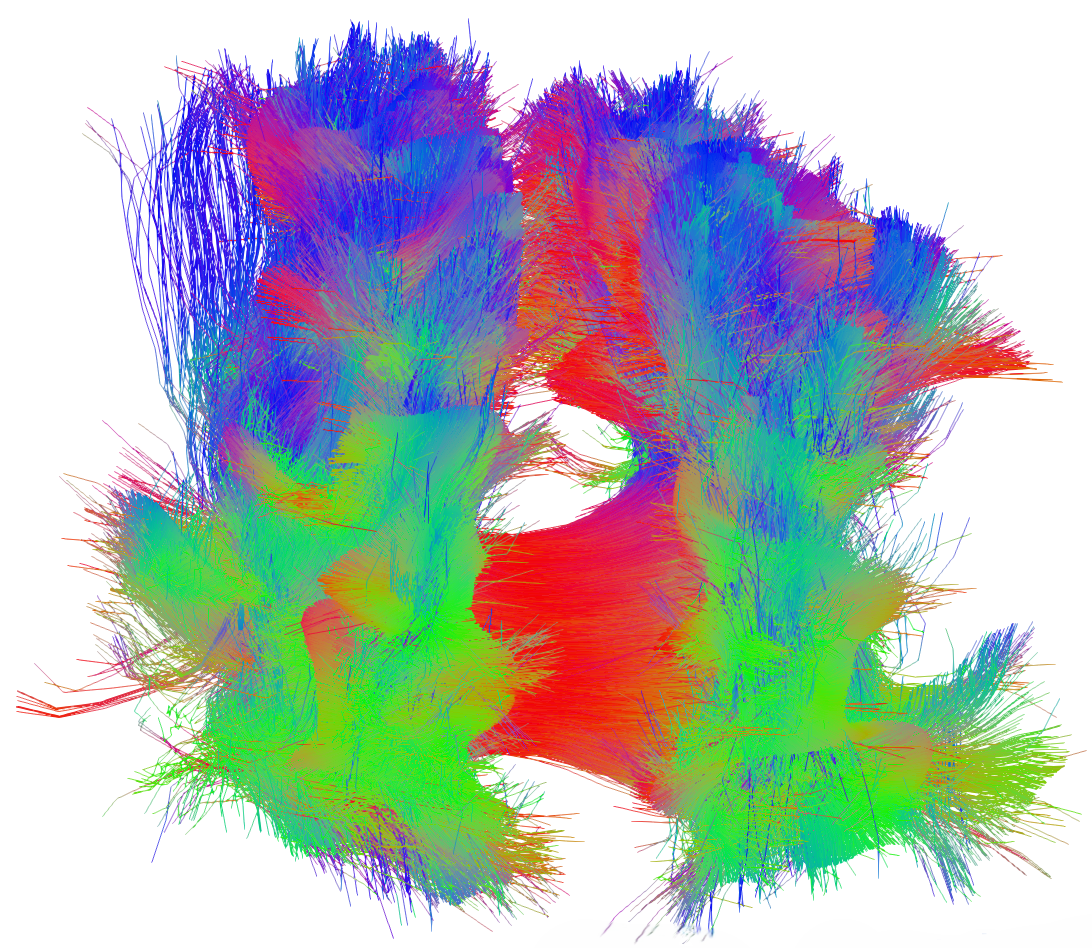}
        \caption*{GT \\ Coronal view}
    \end{subfigure}
    \begin{subfigure}[t]{0.2\textwidth}
        \centering
        \includegraphics[width=\textwidth]{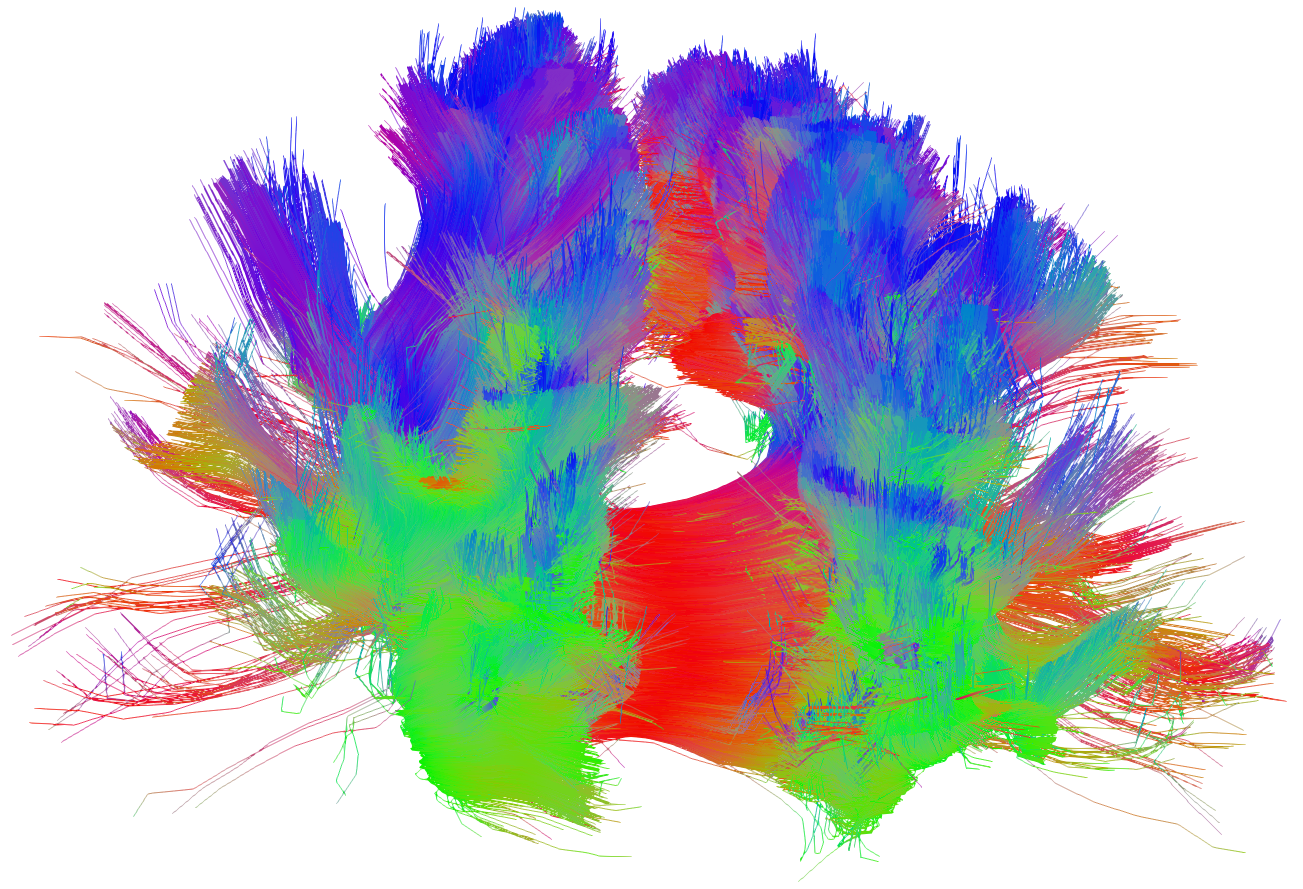}
        \caption*{Ours \\ Coronal view}
    \end{subfigure}
    \hfill
    \begin{subfigure}[t]{0.15\textwidth}
        \centering
        \vspace*{-1.5cm}
        \parbox[c][\textwidth][t]{\textwidth}{%
            \centering
            \textbf{Corpus Callosum (Frontal)}
        }
    \end{subfigure}

    \caption{Visual comparison of tractography outputs from TractoInferno (GT) and TractoTransformer (TT) across four regions. Each row shows a different tract, with matched views from both models.}
    \label{fig:unified_tractography_comparison}
\end{figure*}

To evaluate our method on real (in vivo) \ac{dmri} data, we used the TractoInferno dataset. Due to computational limits, we trained on ten subjects, validated on two, and evaluated on one test subject.
The model architecture and training setup matched those used for the ISMRM dataset, with the key difference being multi-subject training instead of a single brain. Training took approximately seven days. Figure~\ref{fig:unified_tractography_comparison} shows whole-brain and bundle-level tractography outputs from our model alongside ground truth labels for the test subject (sub-1019), demonstrating TractoTransformer’s ability to generalize across subjects and accurately reconstruct complex fiber pathways.


\section{Broader Impact}
\label{impact}
Diffusion MRI tractography is central to understanding brain connectivity, with applications in neuroscience, neurodevelopment, and presurgical planning. Our proposed method, TractoTransformer, improves tract reconstruction by modeling white matter pathways as sequential processes conditioned on anatomical and historical context. This approach aims to reduce false positives and enhance anatomical plausibility—two key challenges in tractography.

Potential societal benefits include advancing neuroscience research and supporting safer, more precise neurosurgical interventions. Improved tractography may also aid early diagnosis and monitoring of neurological disorders, where white matter changes are often early indicators.

Nonetheless, important risks must be considered. Visual plausibility does not imply anatomical accuracy, and overreliance on model outputs—especially in clinical settings—could lead to misinterpretation. Moreover, training on curated datasets like ISMRM and TractoInferno may introduce biases related to acquisition protocols, demographics, or labeling methods, limiting generalization to underrepresented or pathological populations.

To mitigate these risks, we stress that our model is intended as a research tool, not a clinical diagnostic system. Interpretability, robust validation on out-of-distribution data, and collaboration with medical professionals are essential before clinical deployment.


\section{Limitations}
\label{limitations}
Incorporating a \ac{cnn3d} layer significantly increases memory requirements during training, as each voxel in a streamline requires access to \ac{dwi} data from its 26 neighbors (with a kernel size of \(3 \times 3 \times 3\)). In practice, this necessitates loading full-brain \ac{dwi} volumes into memory when training on streamlines from those brains, limiting subject diversity per batch. On our hardware, we were restricted to ten TractoInferno subjects in memory simultaneously.

Although our model achieves strong tractography performance, full-brain inference on a single subject takes approximately 2.5 hours on a single V100 GPU, and can be reduced to roughly 43 minutes by parallelizing across 4 GPUs. To facilitate large-scale applications, future work may further address this through model distillation, batching optimizations, or advanced inference techniques such as speculative decoding.


\section{Conclusions}
\label{conclusions}

We introduced TractoTransformer, a hybrid CNN-Transformer framework for diffusion MRI tractography that leverages both local microstructural context and sequential trajectory modeling. By producing conditional \acp{fodf} at each point along a streamline, our model provides anatomically plausible reconstructions. Quantitative results on the ISMRM dataset and qualitative analysis on TractoInferno validate the efficacy of the method.






\newpage

\bibliographystyle{unsrt}
\bibliography{main}


\appendix



\end{document}